%% file: main.tex
\documentclass[lettersize,journal]{IEEEtran}
\usepackage{amsmath,amsfonts}
\usepackage{amssymb}
\usepackage{algorithmic}
\usepackage{algorithm}
\usepackage{array}
\usepackage{booktabs}
\usepackage{dsfont}
\usepackage[caption=false,font=normalsize,labelfont=sf,textfont=sf]{subfig}
\usepackage{textcomp}
\usepackage{stfloats}
\usepackage{url}
\usepackage{verbatim}
\usepackage{graphicx}
\usepackage{cite}
\usepackage{xspace}
\newcommand{\ie}{{\emph{i.e.}}\xspace}
\newcommand{\eg}{{\emph{e.g.}}\xspace}

\newcommand{\etal}{{\emph{et al.}}\xspace}
\usepackage[pagebackref=false,breaklinks=true,colorlinks,citecolor=black,linkcolor=red,bookmarks=false]{hyperref}
\newcommand{\thickhline}{%
    \noalign {\ifnum 0=`}\fi \hrule height 1pt
    \futurelet \reserved@a \@xhline
}
\usepackage{multirow}
\usepackage{color}

\newcommand{\cmark}{\ding{51}\xspace}%
\newcommand{\xmarkg}{\textcolor{lightgray}{\ding{55}}\xspace}%

\usepackage{bbding}
\usepackage{pifont}
\usepackage{color,xcolor}
\usepackage{colortbl}
\usepackage{stfloats}
\usepackage{wasysym}
\usepackage{xspace}
\usepackage{bbm}
\usepackage{makecell}
\definecolor{myorange}{RGB}{255,72,3}
\definecolor{mygray}{gray}{.85}
\definecolor{mygray1}{gray}{.7}
\definecolor{mygray2}{gray}{.93}
\definecolor{mygray3}{gray}{.98}
\newcommand{\ours}{RGANet\xspace}
\begin{document}
\title{Region Generation and Assessment Network for\\ Occluded Person Re-Identification}
\author{Shuting He,
        Weihua Chen,
        Kai Wang,
        Hao Luo,
        Fan Wang,
        Wei Jiang,
        Henghui Ding
}

\markboth{}%
{Shell \MakeLowercase{\textit{et al.}}: A Sample Article Using IEEEtran.cls for IEEE Journals}

\maketitle
\begin{abstract}
Person Re-identification (ReID) plays a more and more crucial role in recent years with a wide range of applications. Existing ReID methods are suffering from the challenges of misalignment and occlusions, which degrade the performance dramatically. Most methods tackle such challenges by utilizing external tools to locate body parts or exploiting matching strategies. Nevertheless, the inevitable domain gap between the datasets utilized for external tools and the ReID datasets and the complicated matching process make these methods unreliable and sensitive to noises.
In this paper, we propose a \textbf{R}egion \textbf{G}eneration and \textbf{A}ssessment \textbf{Net}work (\textbf{RGANet}) to effectively and efficiently detect the human body regions and highlight the important regions. 
In the proposed \ours, we first devise a Region Generation Module (RGM) which utilizes the pre-trained CLIP to locate the human body regions using semantic prototypes extracted from text descriptions. Learnable prompt is designed to eliminate domain gap between CLIP datasets and ReID datasets. Then, to measure the importance of each generated region, we introduce a Region Assessment Module (RAM) that assigns confidence scores to different regions and reduces the negative impact of the occlusion regions by lower scores. The RAM consists of a discrimination-aware indicator and an invariance-aware indicator, where the former indicates the capability to distinguish from different identities and the latter represents consistency among the images of the same class of human body regions. Extensive experimental results for six widely-used benchmarks including three tasks (occluded, partial, and holistic) demonstrate the superiority of \ours against state-of-the-art methods.

\end{abstract}


\input{1_introduction}

\input{2_relatedworks}

\input{3_methods}

\input{4_experiments}

\input{5_conclusion}

\ifCLASSOPTIONcaptionsoff
  \newpage
\fi

{\small
\bibliographystyle{IEEEtran}
\bibliography{reference}
}

\vfill

\end{document}

%% file: 1_introduction.tex
\section{Introduction}\label{sec:introduction}
\begin{figure}[t]
	\includegraphics[width=0.49\textwidth]{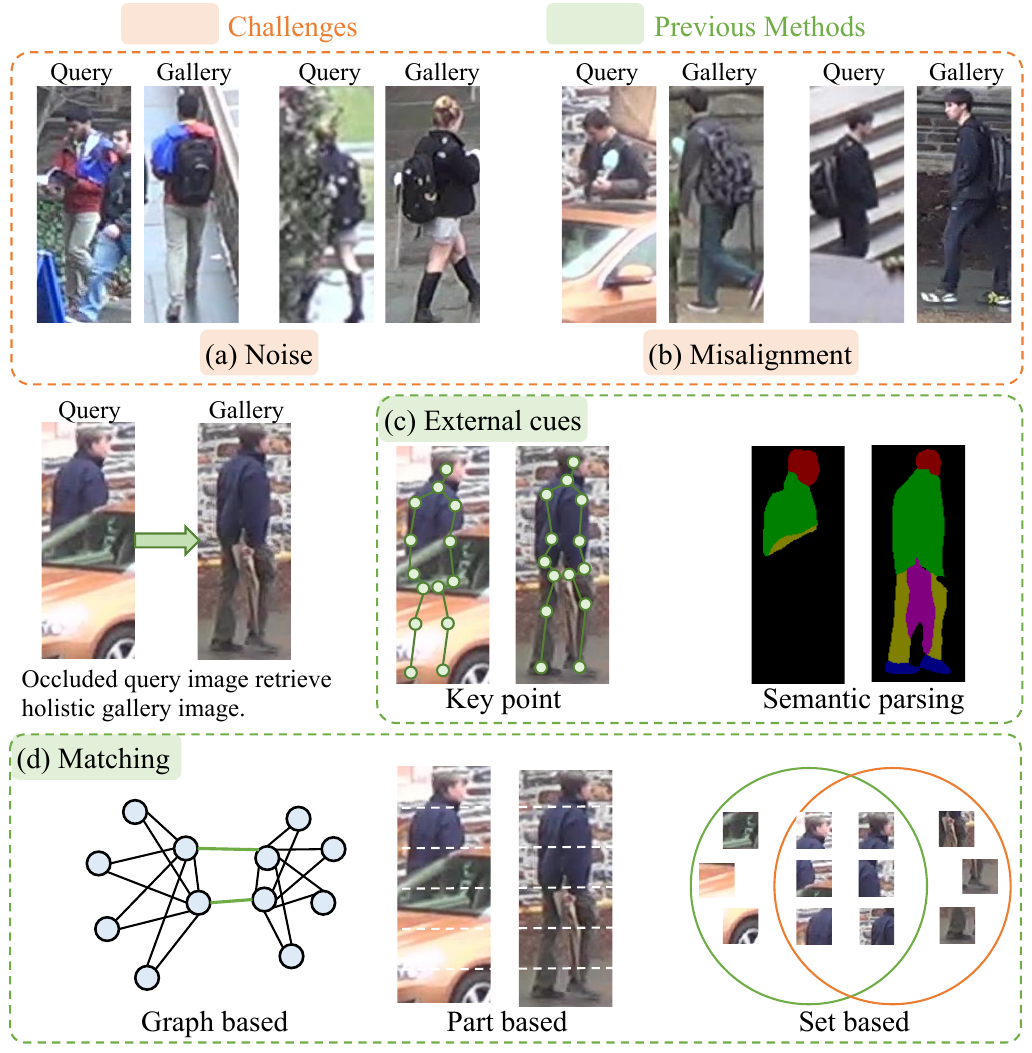}
	\caption{
The upper block illustrates the two key challenges in occluded ReID. (a) Noises caused by occlusions interfere the feature extraction. (b) The occluded image containing only part of the human body results in spatial misalignment. The lower block shows two mainstreams to solve the challenges. (c) External cues are utilized to detect and align body parts. (d) Matching based methods establish alignment relationships between feature patches obtained by specific rules.
 }
	\label{fig:motivation}
\end{figure}

\IEEEPARstart{P}{erson} re-identification (ReID) aims to match images of the target person across different times, locations, and camera views. Holistic person ReID community has witnessed significant improvement by many advanced deep learning algorithms \cite{PCB,MGN,RGA,triplet_loss,SNR,transreid} and well-annotated large-scale datasets \cite{Market1501,DukeMTMC-reID,CUHK03-2,PGFA,partial_iLIDS,partial-REID,COCAS}. 
However, person ReID is still challenging due to the difficulty in extracting robust feature representations under the occlusion scenarios.  
As humans are occluded by clutters and obstacles commonly in practical surveillance systems, occluded person ReID~\cite{zhuo2018occluded,PGFA} deserves further study owing to its great value in real-world scenarios.

Compared with holistic person ReID, there exist two critical issues for occluded person ReID: (1) Occlusions (\eg, cars, trees, boards, persons) bring various noises which interference the extraction of discriminative feature, as shown in \figurename~\ref{fig:motivation}(a). (2) The occluded image contains only part of the human body, leading to the issue of spatial misalignment, as shown in \figurename~\ref{fig:motivation}(b).
Most existing works tackle such challenges by locating the non-occluded body parts and aligning these body parts. These approaches can be categorized into two mainstreams, \ie, external cues based methods and matching based methods.

External cues based methods utilize some off-the-shelf models, \eg, semantic parsing~\cite{SPReID,MGCAM,GASM,CCL,BFP,MSDPA} or key point models~\cite{PGFA,PVPM,HOReID,PGFLKD,GASM,PFD}, to facilitate detecting and aligning the body part, as shown in \figurename~\ref{fig:motivation}(c). 
They generally use external cues to divide the human body into several informative parts, \eg ``\texttt{head}'', ``\texttt{upper body}'', ``\texttt{lower body}'', and ``\texttt{foot}'' from top to bottom.
However,
the inevitable domain gap between the images utilized for training external tools and the ReID datasets significantly poses the obstacle of obtaining an effective and efficient model. 
Matching based strategies~\cite{set_matching,HOReID,PCB,MGN,ISP,DSR,SFR} aim to construct the alignment relationship between local features patches obtained by specific rules as illustrated in \figurename~\ref{fig:motivation}(d).
Nevertheless, complicated matching algorithms lead to high computational costs during the inference stage. Especially, the computational complexity is overwhelming when handling large-scale datasets and matching methods may collapse in severely occluded scenarios.

In this work, we propose an efficient and effective \textbf{R}egion \textbf{G}eneration and \textbf{A}ssessment \textbf{Net}work (\textbf{RGANet}) that enables the network to adaptively locate human body parts and introduce confidence score to guide the network in training and matching. 
We introduce text clues into human part detection and propose a Region Generation Module (RGM) to produce the human body regions using semantic prototypes by textual descriptions. To this end, we utilize the large-scale pre-trained vision-language models, \eg, CLIP~\cite{clip} and ALIGN~\cite{jia2021scaling}, to project semantic words (names of human body regions) to vision-aligned textual prototypes. The large-scale pre-trained vision-language models have strong capability of matching visual and textual features with great generalization ability towards open-world vocabulary.
Therefore, each vision-aligned textual prototype represents the corresponding class and is discriminative enough to distinguish each other.
In such a way, we convert the problem of partitioning each image properly into classifying each pixel of the image to a specific category from ``\texttt{head}'', ``\texttt{upper body}'', ``\texttt{lower body}'', and ``\texttt{foot}''. 
CLIP, being an external model, also exists domain gap with ReID datasets. To address this domain gap, we employ prompt learning to introduce additional learnable context tokens that are trained on the ReID datasets. This process enables it to adapt and align with the specific characteristics of the ReID task. Therefore, the domain gap between the dataset utilized for CLIP and the ReID datasets is effectively eliminated.
We feed these category names into our designed prompt and produce the prototypes by CLIP's pre-trained text encoder. Then, each pixel of the ReID image is labeled by the class-specific prototypes nearest to its feature representation, by which the image feature map is segmented into different regions. Such regions are adaptively generated under the guidance of semantic clues and training objectives, and thus can better capture informative regions than fixed external cues based methods. 
Moreover, once the training is completed, the textual prototypes optimized after the training stage can be directly used for testing, rather than having to extract them again like external tools methods.
By generating these human body regions, we can not only focus more on informative locations that are useful to differentiate different person identities apart, but also suppress background and occluded parts that are detrimental to the ReID performance.

Then, we propose a Region Assessment Module (RAM) to highlight the more importance regions.
The clues contained in these detected regions show different importance in ReID and some occlusion part may have negative impact to the performance. 
For example, when a person's feet are occluded, the detected foot region cannot provide useful information for re-identifying this person. Besides, sometimes it is unavoidable to include some noise of the occlusion or background in the generated informative regions, which will interfere with the learning of the network.
In the field of object detection, 
the credibility of each bounding box will be determined by a confidence score.
Inspired by it, the Region Assessment Module (RAM) is introduced to evaluate the quality of the generated regions and assign different confidence scores to these regions, which greatly helps to reduce the negative impact of the occlusion part. The proposed RAM consists of a discrimination-aware indicator and an invariance-aware indicator, which are used to measure the discrimination and invariance of the features of our generated regions, respectively. 
The discrimination-aware indicator uses a self-learning manner to measure the discrimination capability, \ie, whether this region can help to distinguish different identities.
The invariance-aware indicator aims to learn the invariant information embedded in all images from the same category of the human body regions, \ie, what is consistent among the images of the same category of the human body regions.
Combining these two complementary scores, the confidence score is obtained which is employed in the loss function for robust ReID feature learning and matching. With the help of RAM, the informative region features are more discriminative and invariant, which significantly enhances the feature representation capability of the regions. 
Finally, the final matching process can be performed by the summation of human body regions distance aggregation weighted by the confidence score.

The main contributions of this paper are summarized as follows:
\begin{itemize}
\setlength\itemsep{0.5em}
\item We propose a \textbf{R}egion \textbf{G}eneration and \textbf{A}ssessment \textbf{Net}work (\textbf{RGANet}) to adaptively locate human body parts using semantic information and select the more informative parts to enhance occluded person ReID.
\item We propose a Region Generation Module (RGM) that generates human part regions using semantic representations of CLIP. Regions are thus endowed with a better discriminative ability to mitigate occlusion interference.

\item To further analyze the generated regions, a Region Assessment Module (RAM) is introduced to measure the importance of these regions, which greatly lessens the adverse impact of occlusion or uninformative regions on the network.

\item The proposed \ours achieves state-of-the-art performance on occluded, partial, and holistic ReID benchmarks including Occluded-DukeMTMC~\cite{PGFA}, Occluded-REID~\cite{zhuo2018occluded}, Partial-REID~\cite{partial-REID}, Partial-iLDIS~\cite{partial_iLIDS}, Market-1501~\cite{Market1501}, and MSMT17~\cite{MSMT17}.
\end{itemize}

%% file: 2_relatedworks.tex
\section{Related Work}

\subsection{Occluded Person Re-Identification}

Compared with holistic person ReID, occluded person ReID is more challenging due to noise interference and spatial misalignment. Basically, existing approaches can be divided into two streams, external cues based methods and matching based approaches.

Previous methods utilize external models such as semantic parsing~\cite{MSDPA,SPReID,DIIM,deepdeblur,MGCAM,GASM}, human pose~\cite{PGFA,PVPM,HOReID,PGFLKD,GASM,PFD} to locate aligned body parts. With the help of extra semantic information, 
such methods are capable of aligning parts precisely and extracting more robust feature representation. 
For example, 
To facilitate probe and gallery feature generation and matching, Miao~\etal~\cite{PGFA} propose a pose-guided feature alignment method (PGFA) making use of the external human pose models.
Gao~\etal~\cite{PVPM} design a pose-guided visible part matching algorithm (PVPM). Under the guidance of pose estimation and graph matching, it jointly extracts features and mines the visibility of parts through attention heatmaps.
HOReID is devised by Wang~\etal~\cite{HOReID} to extract discriminative features and achieve robust alignment via building high-order relation and topology information according to the estimation of key-points.
He~\etal~\cite{GASM} introduce a new matching approach GASM
to combine the pose and semantic mask information to output saliency
heatmap which is utilized to further supervise discriminative feature learning and conduct adaptive spatial matching.
Despite external cues contributing to alleviating the challenges of occluded ReID, the inevitable domain gap between the images used for training external models and the ReID
datasets lead to unreliable external cues and hinder the subsequent ReID process.

Matching based strategies~\cite{HOReID,set_matching,PCB,MGN,ISP,MSCAN,DPL} aim to construct the alignment relationship between local features patches obtained by specific rules and it can be further divided into part-to-part matching~\cite{PCB,MGN,ISP,MSCAN,DPL}, graph base matching~\cite{HOReID}, and set based matching~\cite{set_matching}.
Sun~\etal~\cite{PCB} divide feature maps into several horizontal stripe embeddings and train non-shared classifiers on each one.
Using K-means clustering algorithms, Zhu~\etal~\cite{ISP} detect human body parts and additional personal belongings from the pixel level.
Yao~\etal~\cite{DPL} design the part loss to locate human body parts which enforce the network to extract representations for various parts.
HOReID~\cite{HOReID} and PVPM~\cite{PVPM} both apply graph matching to construct alignment relationship between parts.
Jia~\etal~\cite{set_matching} extract features along the channel dimension by a pattern set capturing one particular visual pattern and utilize Jaccard distance instead of Euclidean distance to compute the similarity between pattern sets. 
These methods extract and align local features through a self-supervision manner without external cues. Although complicated matching algorithms lead to high computational costs during the inference stage. Besides, the predicted result largely depends on the quality of obtained regions which is susceptible to noise.
 Differing from the existing works, our \ours aims to explicitly extract the representation of the target person with the help of pre-trained CLIP and the learnable prompt is designed to eliminate domain gap between CLIP datasets and ReID datasets. Moreover, during inference, we no longer require external tools since we have optimized prototypes for segmenting the feature map. Besides, it is capable of awaring of the confidence of features,
which lessens the interference of irrelevant parts.

\subsection{Visual-Language Pretraining}
Nowadays, Visual-Language pretraining has gained increasing attention and reached superior performance on a large number of multi-modal downstream tasks. 
Several methods~\cite{clip, milnce,simvlm} utilize large-scale data sources and exploit semantic supervision to extract visual representations with text representations. 
MIL-NCE~\cite{milnce} has the capability of solving misalignments problem with narrated videos and obtaining strong video feature representations from noisy large-scale dataset HowTo100M~\cite{howto100m}.
SimVLM~\cite{simvlm} lessens the training complexity through taking advantage of large-scale weak supervision which is trained under a single prefix language modeling function in an end-to-end manner.
Recently, Contrastive Language-Image Pretraining (CLIP)~\cite{clip} has obtained remarkable results in multi-modal zero-shot learning, which indicates cross-modal feature representations can be aligned in a shared embedding space. The remarkable generalization capability of CLIP is owing to large-scale training samples of about 400 million image-text pairs obtained from the Internet.

Recently, some researchers introduces the idea of prompt learning, a prevalent trend in
NLP, to the vision area.
Zhou~\etal~\cite{CoOp} introduce Context Optimization (CoOp) which obtain significant improvements over hand-crafted prompts by simply adding a set of learnable vectors in the prompt. 
To improve generalization ability, Zhou~\etal~\cite{CoCoOp} propose Conditional Context Optimization (CoCoOp), which
extends CoOp through further appending a lightweight neural network aiming to generate an input-conditional token for each image.
Our work is built upon the CLIP and leverages its multi-modal alignment capability for retrieving specific semantic parts.

%% file: 3_methods.tex
\section{Approach}
\begin{figure*}[htp]
	\includegraphics[width=0.996\textwidth]{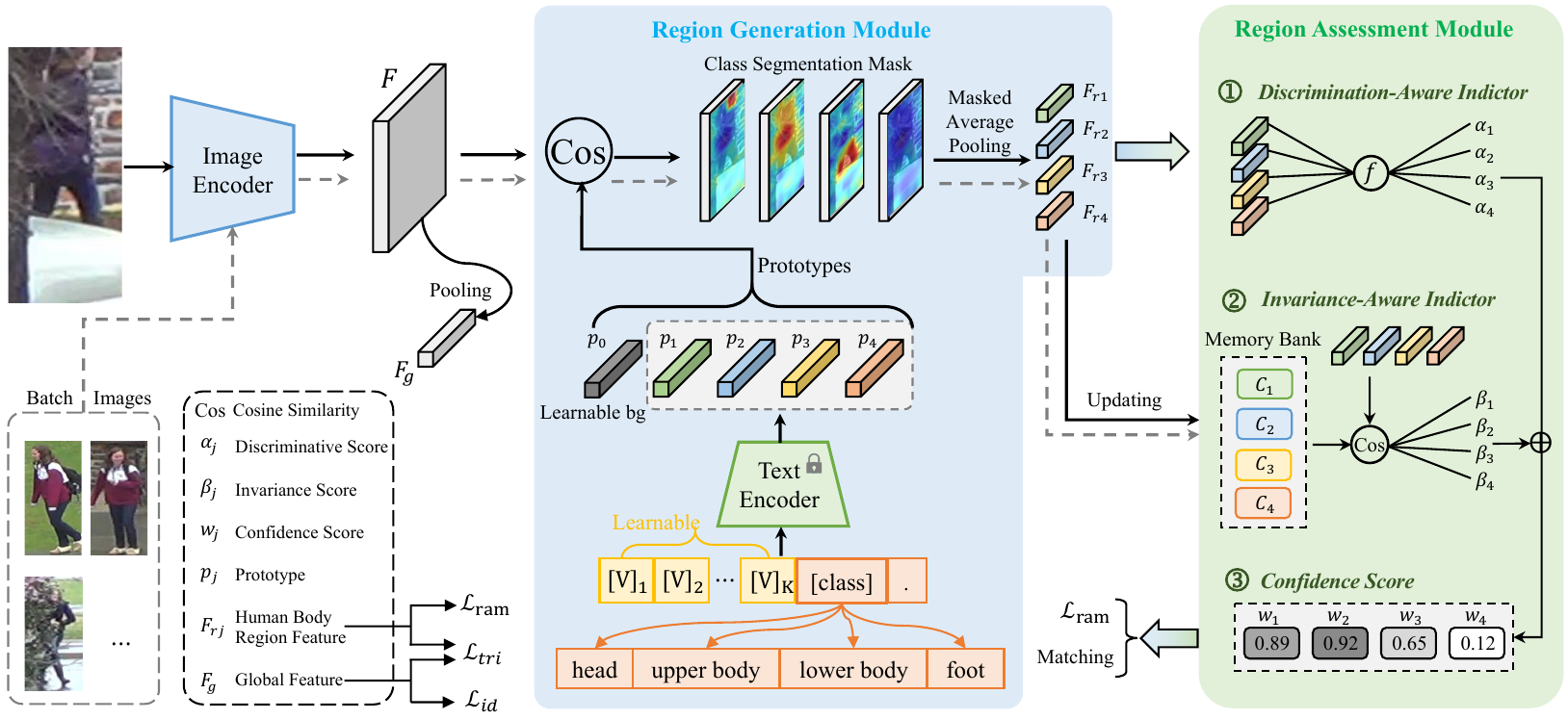}
	\caption{The pipeline of our proposed \textbf{R}egion \textbf{G}eneration and \textbf{A}ssessment \textbf{Net}work (\textbf{RGANet}). It consists of two components: Region Generation Module (middle) and Region Assessment Module (right). The RGM employs CLIP to generate text embeddings as prototypes and then obtain class segmentation mask indicating informative regions by calculating cosine similarity between image features and these textual prototypes. RAM is comprised of the discrimination-aware indicator and the invariance-aware indicator. Discrimination-aware indicator uses $f$, self-attention layer, to measure the discrimination of features for each region. Invariance-aware indicator compares a given human body region feature $F_{j}^r$ with the other same category of the human body regions stored in memory bank $\mathcal{M}$. The gray dotted line indicates that the images  are used to update the memory bank during the training process. Combining these two complementary scores ($\alpha_j$ and $\beta_j$), the confidence score $w_j$ is obtained for further network training and feature matching. }
	\label{fig:pipeline}
\end{figure*}
In this section, we describe the proposed framework of \ours, which is shown in \figurename~\ref{fig:pipeline}. The details of the framework, its two major modules, and its training and testing process will be introduced as below.

\subsection{Overview of \ours}
As shown in \figurename~\ref{fig:pipeline}, we first extract the image feature $F$ by an image encoder.
A global feature $F_g$ is obtained by a Global Average Pooling (GAP) on $F$. 
Region Generation Module (RGM) is applied over the feature map $F$ to locate the position of different classes of human body regions.
Specifically, with the help of CLIP, we first obtain a text-generated prototype,
and class segmentation masks of human parts is generated by labeling each pixel of $F$ as the class of the nearest prototype.
Then the masked average
pooling is utilized to obtain the class-specific region feature $F_{j}^r(j=1, \cdots, N)$.  
To evaluate the confidence of the generated regions, region features are fed into Region Assessment Module (RAM).
RAM consists of two parts, (1)
a self-attention mechanism as discrimination-aware indicator to assign each region with a discrimination score using a fully connected layer followed by a sigmoid function; (2) an invariance-aware indicator utilizing a memory bank to compare the given region with its aligned regions obtains an invariance score. 

\subsection{Revisiting CLIP}

Contrastive Language-Image Pre-training (CLIP)~\cite{clip} first extracts image and text features by an image encoder and a text encoder, respectively. Then the image features are forced to match with their corresponding text features under a contrastive loss, which enables image features and their paired text features to be aligned accordingly. After pre-training, given a category name and image of an unseen category that has never been seen in the training of CLIP, 
CLIP can well match the textual feature obtained according to the category name and visual feature of the image, showing a great generalization ability to unseen categories.

CLIP shows strong performance in many vision tasks, for example, zero-shot learning~\cite{PADing,ZSS,D2Zero} and referring image segmentation~\cite{GRES,VLTPAMI}. A direct way to apply CLIP for zero-shot learning is adopting text embeddings generated by the text encoder of CLIP as weights of a prototype to label visual features into a specific class, \ie, $<T_i, I_i>$,
where $I_i$ is image feature encoded by CLIP’s image encoder, $T_i$ is prototype weights from CLIP's text encoder, and $<\cdot,\cdot>$ represents calculating cosine similarity. Inspired by it, in this work, we take advantage of the strong generalization ability of CLIP and input category names of human body regions into a text encoder of CLIP, generating semantic embeddings of the human body for classifying informative regions.

\subsection{Region Generation Module}

To locate the informative regions and suppress the influence of occlusions/background, we introduce a Region Generation Module (RGM) that aims to adaptively capture more discriminative and well-aligned regions. The architecture of our proposed Region Generation Module is shown in the middle part of \figurename~\ref{fig:pipeline}. In the proposed RGM, we first employ CLIP to generate text embeddings as prototypes and then obtain class segmentation masks indicating informative regions by calculating cosine similarity between image features and these textual prototypes.

\textbf{Prototype Extraction}.
We assume the class names set has $N$ categories including ``\texttt{head}'', ``\texttt{upper body}'', ``\texttt{lower body}'', ``\texttt{foot}''.
The class name from the class name set is placed into a learnable prompt ( introduced in \textbf{Prompt Design}) and then fed into the CLIP text encoder.
Then we can obtain $N$ text embeddings, denoted as $\mathcal{P} = \{p_j\in \mathbb{R}^d |j=1,\cdots,N\}$.
In our ReID task, we also need a ``\texttt{background}'' category that indicates pixels do not belong to any human part classes.
For the ``\texttt{background}'' category, 
one class name is insufficient to describe it.
As such, we add an extra learnable embedding $p_0\in \mathbb{R}^d $ for ``\texttt{background}'' class to represent the comprehensive analysis.
Since we have obtained optimized prototypes for segmenting the feature map, we do not need to extract prototypes again during inference which is superior to the external tools based approaches.

\textbf{Class Segmentation Mask}.
Meantime, we get the image feature map by our image encoder, denoted as $F\in \mathbb{R}^{d\times H \times W}$, where $d$, $H$, $W$ are channel numbers, height, and width of the feature map, respectively. $F$ has the same channel number as these textual embeddings. Each textual embedding contains category-related discriminative clues and is employed to measure how similar the pixel features in $F$ are to the text embedding. To get class segmentation masks $\mathcal{S}_{j} \in \mathbb{R}^{H\times W}(j=1,\cdots,N)$ for each of the human body regions categories, we compute the cosine similarity between image feature $F$ and each of the textual embeddings and apply a softmax over the similarity maps to normalize them, \ie,
\begin{equation}
\label{eqn:prob_map}
    \mathcal{S}_{j}^{(x,y)} = \cfrac{\exp(\gamma <F^{(x, y)}, p_j>)}{\sum_{p_k\in\ \{\mathcal{P} \cup p_0\}}\exp(\gamma <F^{(x, y)}, p_k>)},
\end{equation}
where $(x,y)$ denotes spatial position of pixels, $\gamma$ is an amplification factor, $<a,b>$ indicates calculating cosine similarity between $a$ and $b$. Each position in $S_j$ denotes the similarity between this position's visual feature with $j$-th textual embedding $p_j$. 
``\texttt{background}'' is used to suppress the similarity on pixels that belong to none of the human body regions. Then, we conduct masked average pooling over image feature $F$ to obtain the global representation of each human body region feature $F_{j}^r$,
\begin{equation}
\label{eqn:region_feature}
    F_{j}^r = \cfrac{\sum_{x,y}S_j^{(x,y)}F^{(x,y)}}{\sum_{x,y}S_j^{(x,y)}},
\end{equation}
where $F_{j}^r \in \mathbb{R}^{d}(j=1,\cdots,N)$.
The proposed Region Generation Module (RGM) detects informative human body regions, by measuring feature similarity between visual features and textual features of human body regions classes provided by a pre-trained CLIP model. The generated regions emphasize body part-related positions while suppressing unrelated ones, as shown in \figurename~\ref{fig:pipeline}. With the generated human body-related regions, we can focus more on extracting identification-related clues from these semantic regions while avoiding being affected by noisy information.

\textbf{Prompt Design}. The vanilla CLIP is not designed for person ReID and we explore designing feasible text prompts for person ReID.
A direct way to use CLIP is by utilizing the \texttt{hand-crafted prompts},
while prompts given by CLIP are specifically suitable for image classification and may have an adverse effect on our task.
Consequently, we experiment with different prompts designs to find the most useful one for our RGM, which is ``a [CLASS] part of a person''.
Recently, \texttt{learning-based prompt}~\cite{CoOp,CoCoOp} shows great promise for adapting the CLIP by prompt design on different downstream tasks. 
Inspired by it, we introduce learnable vectors to model prompt context for occluded ReID task. Specifically, a generalized prompt can be expressed as 
$[\text{V}]_1 [\text{V}]_2 \hdots [\text{V}]_K [\text{CLASS}]$
where each $[\text{V}]_k$ ($k\!\in\!\{1, \hdots, K\}$) is a vector set as learnable parameters which have the same dimensions with word embeddings (\ie, 512 for CLIP), and $K$ is the number of learnable context vectors.
An additional training process is indispensable to optimize these context parameters.

To supervise the prompt learning, we perform the segmentation loss $\mathcal{L}_{\text{seg}}$ as follows:
\begin{equation}
\label{eqn: loss_seg}
    \mathcal{L}_{\text{seg}} = \sum_{x,y}\sum_{j=1}^N \mathds{1}[\hat{\mathcal{S}}^{(x,y)}=j]\log  \mathcal{S}_{j}^{(x,y)},
\end{equation}
where $\hat{\mathcal{S}}$ is the pseudo ground truth segmentation mask generated from the previous work~\cite{wang2020vitaa}. 
$\mathds{1}[\cdot]$ is an indicator math function, if the argument is true it output $1$, or $0$ otherwise.
The incorporation of this learnable prompt effectively adapts CLIP to the ReID context.
It is worth noting that hand-crafted prompts do not require additional training procedures or external pseudo-segmentation masks and still achieve decent results (please kindly refer to ablation stduy in TABLE \ref{tab:prompt_design} for details).

\subsection{Region Assessment Module}
The clues contained in these body part regions play different roles in ReID. Besides, when facing occlusion, the corresponding regions contain less useful information and may bring noise. If these occluded regions are fed into the training of the network as the informative regions, it will interfere with learning discriminative features, because the network may regard the same occasion as the signal of the same person wrongly. In the field of object detection, a confidence score will be assigned to each bounding box to estimate its credibility.
Hence, inspired by it, we propose a Region Assessment Module (RAM) to give them different confidence scores so that they make different contributions to the network training and matching. 
The confidence score assigned to each region here is designed by taking these two aspects into consideration, \ie, discrimination among different classes, and invariance within the same classes\footnote{Here classes refer to the human body region classes.}. 

\textbf{Discrimination-Aware~Indicator.} This indicator is designed to capture the region's contributions to each sample. It is expected that the discrimination-aware indicator assigns higher scores to the more discriminative regions and lower scores to the background or occluded regions. 

The discrimination-aware indicator takes $F_{j}^r$ as input and outputs a discrimination score for each region. Specifically, it is composed of a fully connected (FC) layer and followed by a sigmoid activation function, which can be expressed as,
\begin{equation}
\alpha_j = \sigma(W_{a}^{\top}F_{j}^r),
\end{equation}
where $\alpha_j$ is the discrimination score of the $j$-th region, the parameters of the FC layer are defined by $W_a \in \mathbb{R}^{d\times 1}$ and the sigmoid function is defined by $\sigma$. The discrimination-aware indicator pushes the features to learn the inherent relations of images from different identities and measures the discrimination ability of each region. It is worth noting that the parameters of the indicator are shared among different regions.

\textbf{Invariance-Aware Indicator.}
The invariance-aware indicator makes sure that the features of samples within same category of the human body regions would explore the invariant characteristic shared within these regions. 

We introduce a memory bank $\mathcal{M}$ to collect invariance characteristics, inspired by \cite{spcl}. 
The key process of $\mathcal{M}$ generation includes memory initialization and memory update. Unlike~\cite{spcl}, we initialize the memory with the class centers instead of ID centers in the training set. We utilize average features of the same category of the human body regions to act as class centers. In the early stage of network training, the features from RGM may not be extracted correctly, so we use the fixed-stripe feature to initialize the memory.
It is worth noting that a forward step is executed at the beginning of the training process to initialize the class centers, which are then constantly updated as training progresses.
The mean of the extracted features with the same category of the human body regions $j$ are utilized to update the $j$-th center $C_j$ in the mini-batch
\begin{equation}
C_j = m_u C_j + (1-m_u)\frac{1}{|B_j|}\sum_{F_{j}^r \in {B_j}} F_{j}^r,
\end{equation}
where $B_j$ represents the human body region features involved with class $j$ in the mini-batch, $m_u$ denote the momentum updating rate, $F_{j}^r$ is the region feature produced by RGM.
And we select regions with $\alpha_j$ greater than 0.85 to update the memory bank which ensures that the features in the memory bank are high quality as possible.

Given a region tensor $F_{j}^r$, we take out the aligned class center $C_{j}$ and compute the cosine similarity between $F_{j}^r$ and $C_j$ to get invariance score $\beta_j$ as:

\begin{equation}
    \beta_j = \cfrac{\exp( <F_{j}^r, C_j>)}{\sum_{j=1}^N\exp(<F_{j}^r, C_j>)},
\end{equation}
where $\beta_j$ denotes the relevance between the given region feature $F_{j}^r$ and $j$-th class center $C_{j}$. $<a,b>$ indicates calculating cosine similarity between $a$ and $b$.
With the help of an invariance-aware indicator, when there is a region that is full of occlusions, it will give a small score to suppress its impact on the network after comparing it with a common well-generated region in the memory bank.

\textbf{Combination of the Indicators.}
Finally, the $\alpha_j$ obtained by discrimination-aware indicator and the $\beta_j$ collected by invariance-aware indicator are summed together to get the final comprehensive confidence score $w_j$. 
A softmax function is then used to normalize the $w_j$ to obtain the final confidence score.
We apply this confidence score $w_j$ to the loss function and in the later feature matching in Eq.~(\ref{eqn:feature_matching}).
With the confidence score $w_j$, we multiply it by the cross-entropy loss of ReID. The loss function is described as: 
\begin{equation}
\mathcal{L}_\text{ram} = -w_j log(softmax(W_b^\top F_{j}^r)),
\end{equation}
where $W_b$ is the classifier for $j$-th region. The $\mathcal{L}_\text{ram}$ has a positive correlation with the $w_j$.
When the quality of the given feature is not satisfactory, the network will produce a small $w_j$. Therefore, the impact on network training will be small, and vice versa.
In summary, RAM combines two complementary aspects: discrimination and invariance to perceive the contribution of the generated regions and provides a comprehensive awareness of each given region.

\subsection{Training and Testing Process}
In the training phase, we calculate cross-entropy and triplet losses\cite{triplet_loss} for both region's features and global features. The overall objective function is formulated as:
\begin{equation}
\resizebox{.9\hsize}{!}{$
\mathcal{L}_{total} = \sum_{j}\mathcal{L}_\text{ram}(F_{j}^r)+\sum_{j}\mathcal{L}_{tri}(F_{j}^r)+\mathcal{L}_{id}(F_g)+\mathcal{L}_{tri}(F_g),
$}
\end{equation}
where $\mathcal{L}_{id}$, $\mathcal{L}_{tri}$ denotes the cross-entropy loss, triplet losses, respectively. 

In the testing phase, inspired by the \cite{PGFA}, the distance $d_j$ of the $j$-th region between query and gallery images is:
\begin{equation}
d_j = D(F^{r,q}_{j},F^{r,g}_{j}) (j = 1,\ldots,N),
\end{equation}
where $D(\cdot, \cdot)$ represents the cosine distance. 
$F^{r,q}_{j}$, $F^{r,g}_{j}$ represents the $j$-th region feature of the query and gallery image, respectively. Analogously, the distance between global features is given by: $d_g = D(F^q_g,F^g_g)$. Finally, the distance $d$ can be calculated as follows:
\begin{equation}
d =\frac{\sum_{j=1}^{N}(w^q_j \cdot w^g_j)d_j+d_g}{\sum_{j=1}^{N}(w^q_j \cdot w^g_j)+1}.
\label{eqn:feature_matching}
\end{equation}

%% file: 4_experiments.tex
\section{Experiments}
\subsection{Datasets}

\textbf{Occluded person ReID datasets}. We evaluate the effectiveness of our method in occlusion scenarios. \textbf{Occluded-DukeMTMC}~\cite{PGFA} contains 15,618 training images, 17,661 gallery images, and 2,210 occluded query images. Occluded-DukeMTMC contains pictures from DukeMTMC-reID~\cite{DukeMTMC-reID}, while training, query, and gallery contain 9\%, 100\%, and 10\% occluded pictures respectively.
\textbf{Occluded-REID}~\cite{zhuo2018occluded} is an occluded person ReID dataset obtained from mobile cameras. A total of 2,000 images of 200 individuals are contained within it.
For each identity, five full-body images and five occluded images are provided, each with a different viewpoint and an occlusion of a different severity.

\textbf{Partial person ReID datasets}. Partial-REID\cite{partial-REID} and Partial-iLDIS\cite{partial_iLIDS} are two widely used datasets for partial ReID task. 
\textbf{Partial-REID} contains 600 images and 60 people, each of which comprises of five holistic images and five partial images.
\textbf{Partial-iLIDS} contains 238 images of 119 people from multiple non-overlapping airport cameras, with non-occluded areas manually cropped. In order to make a fair comparison with other competitive methods, we use Market-1501 as a training set and two partial datasets as test sets~\cite{VPM, PGFA, HOReID, DSR,SFR}.

\textbf{Holistic person ReID datasets}. We also conduct experiments on two widely used holistic person ReID benchmarks Market-1501~\cite{Market1501} and MSMT17~\cite{MSMT17} where few images are occluded. 
In \textbf{Market-1501}, the number of training and testing images is 12,936 and 19,732 respectively, with 1,501 identities in all.
\textbf{MSMT17} dataset contains 126,441 images of 4,101 identities captured from 15 cameras which makes it more challenging. There are 32,621 images of 1,041 identities for training, 93,820 images of 3,060 identities for testing. During inference, 11,659 from 93,820 images are randomly chosen as the query and the other images are viewed as the gallery.

\subsection{Implementation}

\textbf{Experimental details:}
All training images are resized to $256 \times 128$ and augmented with random horizontal flipping, padding with 10 pixels, random cropping, and random erasing. The batch size is set to 64 with 4 images per person. We use ViT pre-trained on CLIP as our backbone. Adam optimizer is employed with the weight decay factor of 0.0005. The learning rate is initialized as 5e-5 and decreased by a factor of 0.1 at 40th and 70th epochs, respectively, and the training is stopped at 120 epochs. 
For the prompt training process, we freeze all the other parameters and just train the learnable context vectors under the guidance of $\mathcal{L}_{\text{seg}}$ loss. We employ the Adam optimizer with a learning rate of 5e-5 for this purpose. After 30 epochs of training, we obtain an optimized prompt context, which serves as our learnable prompt specifically designed for the ReID task.
The number of the generated regions $N$, the amplification factor $\gamma$, the length of learnable prompts $K$, and the momentum updating rate $m_u$ are set to 4, 20, 8, and 0.3, respectively. Category names fed into CLIP text encoder are set to`` \texttt{head}'', `` \texttt{upper body}'', `` \texttt{lower body}'', `` \texttt{foot}''. The influences of these hyper-parameters will be investigated in the following ablation studies. We freeze the CLIP text encoder during both training and inference.
All the experiments are performed with one Nvidia RTX TITAN GPU card using the PyTorch toolbox\footnote{http://pytorch.org}.  

\textbf{Evaluation Protocols.} As most previous methods in ReID community, Cumulative Matching Characteristic (CMC) curves and the mean Average Precision (mAP) are utilized as metrics to estimate the algorithm. All the experimental results are conducted under the setting of a single query.

\renewcommand{\multirowsetup}{\centering}
\begin{table}[t]
  \begin{center}
   \caption{\label{tab:mdl-2}Performance comparison of the occluded ReID problem on the Occluded-DukeMTMC and Occluded-ReID. These SOTA methods are categorized into four groups from top to bottom: holistic ReID based, matching based, external cues based, and transformer based.}
\setlength{\tabcolsep}{3.mm}
\renewcommand{\arraystretch}{1.1}{
  \begin{tabular}{ l|cc|cc}
\Xhline{1.0pt}
\rowcolor{mygray}	 & \multicolumn{2}{c|}{Occluded-Duke} &  
\multicolumn{2}{c}{Occluded-REID}\\
\rowcolor{mygray}\multirow{-2}{*}{Method}& Rank-1 & mAP & Rank-1 & mAP\\
\hline\hline

Part-Aligned~\cite{PAR} 	& 28.8& 20.2 & - & -  \\
PCB~\cite{PCB} 	& 42.6& 33.7 & 41.3 &  38.9  \\
Adver Occluded~\cite{adver_occluded} 	& 44.5 & 32.2 & - & -  \\
\hline
DSR~\cite{DSR} 	& 40.8 & 30.4 & 72.8 &62.8   \\
SFR~\cite{SFR} 	& 42.3 & 32.0& - & -  \\
FRR~\cite{FPR}&- & - & 78.3 & 68.0  \\
MoS~\cite{set_matching} &61.0 & 49.2 & - & -\\
\hline
PVPM~\cite{PVPM} 	& 47.0 & 37.7 & 70.4 &61.2   \\
PGFA~\cite{PGFA} 	& 51.4 & 37.3 & - & -  \\

HOReID~\cite{HOReID} 	& 55.1& 43.8 &80.3& 70.2  \\
GASM~\cite{GASM}& -& - &74.5& 65.6 \\
VAN~\cite{VAN} & 62.2& 46.3 & - &   - \\
OAMN~\cite{OAMN} 	&62.6 &46.1  &-  & -   \\
PGFL-KD~\cite{PGFLKD}&  63.0 &54.1 &80.7& 70.3    \\
\hline
PAT~\cite{PAT} 	& 64.5 & 53.6 &  81.6 & 72.1   \\
DRL-Net~\cite{DRL} 	 & 65.8 &  53.9& - & -  \\
FED~\cite{FED} 	 & 68.1 &  56.4&86.3  & 79.3  \\
MSDPA~\cite{MSDPA} 	& 70.4 & 61.7 & 81.9 & 77.5   \\
FRT~\cite{FRT} 	 & 70.7 &  61.3& 80.4 & 71.0  \\
DPM~\cite{DPM} 	& 71.4 & 61.8 & 85.5 &  79.7  \\
\hline
\hline
\textbf{\ours} (Ours) 	& \textbf{71.6} & \textbf{62.4} & \textbf{86.4} & \textbf{80.0}  \\
\hline
  \end{tabular}}
  \end{center}
\end{table}

\subsection{Comparison with State-of-the-Art Methods}

We compare the proposed \ours with current state-of-the-art methods on all the above-mentioned ReID datasets of different scenarios.

\textbf{Evaluation on Occluded Person ReID Dataset.} To validate the superiority of the \ours, we evaluate the \ours on the Occluded Person ReID datasets and show the comparisons in \tablename~\ref{tab:mdl-2}. SOTA methods are divided into four mainstreams: holistic ReID methods without special design for occlusions (Part-Aligned~\cite{PAR}, PCB~\cite{PCB} and Adver Occluded~\cite{adver_occluded}), occluded ReID methods utilizing matching based approaches (DSR~\cite{DSR}, SFR~\cite{SFR}, FPR~\cite{FPR}, and MoS~\cite{set_matching}), occluded ReID methods depending on external cues (PVPM~\cite{PVPM}, PGFA~\cite{PGFA}, HOReID~\cite{HOReID}, GASM~\cite{GASM}, VAN~\cite{VAN}, OAMN~\cite{OAMN}, and PGFL-KD~\cite{PGFLKD}), and methods based on transformer (PAT~\cite{PAT}, FED~\cite{FED}, DRL-Net~\cite{DRL}, MSDPA~\cite{MSDPA}, FRT~\cite{FRT}, and DPM~\cite{DPM}).
Our \ours outperforms all methods by a large margin, attaining 71.6\%/62.4\% and 86.4\%/80.0\% in terms of Rank1/mAP on Occluded-DukeMTMC and Occluded-ReID, respectively.
For example, PAT~\cite{PAT} exploit transformer to get long range relationships among images to solve occlusions occasions and achieve appealing performance. Our \ours still surpasses it by +8.8\% mAP and +7.9\% mAP on Occluded-DukeMTMC and Occluded-ReID, respectively.

\textbf{Evaluation on Partial Person ReID Dataset.} In Partial ReID, raw images are manually cropped through a bounding box to solve the matching problem. 
As a result, severe distortion, misalignment, and occlusions are inevitable, which increases the difficulty of matching.
To further study our proposed \ours, in \tablename~\ref{tab:partial}, we also report the comparison of the results on two partial ReID dataset. As can be seen, 
our \ours achieves 87.2\%/93.5\% and 77.0\%/87.6\% in terms of Rank-1/Rank-3 on Partial-REID and Partial-iLDIS, respectively.
Note that, our \ours is more practical in real-world scene applications because it does not need any external tools during the inference stage.

\renewcommand{\multirowsetup}{\centering}
\begin{table}[t]
  \begin{center}
   \caption{\label{tab:partial}Performance comparison of the partial ReID problem on two partial datasets, Partial-REID\cite{partial-REID} and Partial-iLDIS\cite{partial_iLIDS}.}
\setlength{\tabcolsep}{3.mm}
\renewcommand{\arraystretch}{1.1}{
  \begin{tabular}{ l|cc|cc}
\Xhline{1.0pt}
     \rowcolor{mygray}	 & \multicolumn{2}{c|}{Partial-REID} &  
     \multicolumn{2}{c}{Partial-iLIDS}\\
    \rowcolor{mygray}\multirow{-2}{*}{Method}	 & Rank-1 & Rank-3 & Rank-1 & Rank-3\\
 \hline
 \hline
DSR\cite{DSR} 	& 50.7 & 70.0 & 58.8 & 67.2  \\
SFR\cite{SFR} 	& 56.9& 78.5 & 63.9 & 74.8  \\
VPM\cite{VPM} 	& 67.7& 81.9 & 65.5 &  74.8  \\
PGFA\cite{PGFA} 	& 68.0 & 80.0 & 69.1 & 80.9  \\
PVPM\cite{PVPM} 	& 78.3 & 89.7 & - & -  \\
FPR\cite{FPR} 	& 81.0 & - & 68.1& -  \\
PGFL-KD~\cite{PGFLKD}&  85.1  & 90.8 & 74.0 & 86.7    \\
HOReID\cite{HOReID} 	& 85.3 & 91.0 & 72.6 & 86.4  \\
OAMN~\cite{OAMN} &  86.0& - & 77.3 & -    \\
\hline
FED~\cite{FED} 	 & 84.6 &  82.3& - & -  \\
MSDPA~\cite{MSDPA} 	 & 86.3 &  93.3& 76.5 & 87.4  \\
\hline
\hline
\textbf{\ours} (Ours) 	& \textbf{87.2}  &\textbf{93.5} &\textbf{77.0}  & \textbf{87.6}   \\
\hline
  \end{tabular}}
  \end{center}
\end{table}

\textbf{Evaluation on Holistic Person ReID Datasets.} 
To further validate the effectiveness of our proposed \ours, we perform experiments on Holistic datasets including Market-1501 and MSMT17. As shown in \tablename~\ref{tab:market1501},
the most recent state-of-the-art methods of ReID can be classified into four groups, namely global feature based models (VCFL~\cite{VCFL}, MVPM~\cite{MVPM}, SFT~\cite{SFT}, DMML~\cite{DMML}, IANet~\cite{IANet}, Circle~\cite{Circle_loss}, and Mos~\cite{set_matching}), part feature based models (DSR~\cite{DSR}, PCB~\cite{PCB}, VPM~\cite{VPM}, and ISP~\cite{ISP}), external cues based methods (MGCAM~\cite{MGCAM}, Pose-transfer~\cite{PoseTransfer}, PSE~\cite{PSE}, SPReID~\cite{SPReID}, PGFA~\cite{PGFA}, AANet~\cite{AANet}, HOReID~\cite{HOReID}, GASM~\cite{GASM}, and PGFL-KD~\cite{PGFLKD}), and transformer based models (PAT~\cite{PAT}, TransReID~\cite{transreid}, DRL-Net~\cite{DRL}, FED~\cite{FED}, MSDPA~\cite{MSDPA}, FRT~\cite{FRT}, PFD~\cite{PFD}, and DPM~\cite{DPM}). Our method achieves state-of-the-art performance with 95.5\%/89.8\% Rank-1/mAP on Market-1501.

As can be seen from \tablename~\ref{tab:msmt17}. On MSMT17, our method surpasses the previous holistic SOTA method TransReID~\cite{transreid} by a large margin. \eg, +1.9\% and +2.9\% in terms of Rank-1 and mAP, respectively.
The above experiments demonstrate that our method is a superior universal method for person ReID, which has strong robustness towards different tasks.

\begin{table}[t]
    \centering
    \caption{Comparison with SOTA methods over Market-1501. The compared methods are categorized into three mainstreams: global feature based, part feature based, external cues based, and transformer based.}
    \label{tab:comparison}
    \setlength{\tabcolsep}{4.5mm}
\renewcommand{\arraystretch}{1.1}{
    \begin{tabular}{l|l|c c}
\Xhline{1.0pt}
        \rowcolor{mygray} &  & \multicolumn{2}{c}{Market-1501} \\
        \cline{3-4}
        \rowcolor{mygray}\multirow{-2}{*}{Methods} & \multirow{-2}{*}{Reference}& Rank-1 & mAP  \\
        \hline 
        \hline
        VCFL~\cite{VCFL} &ICCV’19 & 89.3 & 74.5 \\
        MVPM~\cite{MVPM} &ICCV’19 & 91.4 & 80.5  \\
        SFT~\cite{SFT} &ICCV’19 & 93.4 & 82.7  \\
        DMML~\cite{DMML} &ICCV’19 & 93.5 & 81.6  \\
        IANet~\cite{IANet}& CVPR’19 & 94.4 & 83.1  \\
        Circle~\cite{Circle_loss} &CVPR’20 & 94.2 & 84.9  \\
       MoS~\cite{set_matching} &AAAI’21 & 94.7 & 86.8  \\
        \hline
    DSR~\cite{DSR} &CVPR’18 & 83.6 & 64.3  \\
    PCB~\cite{PCB}&ECCV’18 & 92.3 & 77.4  \\
    VPM~\cite{VPM} &CVPR’19 & 93.0 & 80.8  \\
    PCB+RPP~\cite{PCB} &ECCV’18 & 93.8 & 81.6 \\
    ISP~\cite{ISP} &ECCV’20 & 95.3 & 88.6  \\
        \hline
        MGCAM~\cite{MGCAM} &CVPR’18 & 83.8 & 74.3  \\
        Pose-transfer~\cite{PoseTransfer} &CVPR’18 & 87.7 & 68.9  \\
        PSE~\cite{PSE} &CVPR’18 & 87.7 & 69.0 \\
        SPReID~\cite{SPReID} &CVPR’18 & 92.5 & 81.3  \\
        PGFA~\cite{PGFA} &ICCV’19 & 91.2 & 76.8  \\
        AANet~\cite{AANet} &CVPR’19 & 93.9 & 82.5  \\
        HOReID~\cite{HOReID} &CVPR’20 & 94.2 & 84.9  \\
        GASM~\cite{GASM} &CVPR’20 & 94.2 & 84.9  \\
        PGFL-KD~\cite{PGFLKD} &CVPR’20 & 94.2 & 84.9 \\
        \hline
        PAT~\cite{PAT} &CVPR’21 & 94.2 & 84.9  \\
        TransReID~\cite{transreid} &ICCV’21 & 95.2& 88.9 \\
        DRL-Net~\cite{DRL} &TMM’22 & 94.2 & 84.9 \\
        FED~\cite{FED} &CVPR’22 & 95.0 & 86.3 \\
        MSDPA~\cite{MSDPA} &ACM MM’22 & 95.4 & 89.5 \\
        FRT~\cite{FRT} &TIP’22 & 95.5 & 88.1 \\
        PFD~\cite{PFD} &AAAI’22 & 95.5 & 89.7 \\
        DPM~\cite{DPM} &ACM MM’22 & 95.5 & 89.7 \\
        \hline\hline
        \textbf{\ours} (Ours) & IEEE TIFS& \textbf{95.5} & \textbf{89.8} \\
        \hline
    \end{tabular}}
    \label{tab:market1501}
\end{table}

\begin{table}[t]
\setlength{\tabcolsep}{12pt}
    \centering
        \caption{Performance comparison with state-of-the-art models on MSMT17.}
\renewcommand{\arraystretch}{1.1}{
    \begin{tabular}{l|l|c c}
\Xhline{1.0pt}
\rowcolor{mygray} & & \multicolumn{2}{c}{MSMT17} \\
        \cline{3-4}
\rowcolor{mygray} \multirow{-2}{*}{Methods}        & \multirow{-2}{*}{Reference} & Rank-1 & mAP  \\
    \hline\hline
    MVPM~\cite{MVPM} &ICCV’19    &71.3   & 46.3      \\
    SFT~\cite{SFT} &ICCV’19     &73.6   & 47.6      \\
    IANet~\cite{IANet} &CVPR’19   &75.5   & 46.8      \\
    DG-Net~\cite{DGNet} &CVPR’19  &77.2   & 52.3      \\
    OSNet~\cite{OSNet} &ICCV’19   &78.7   & 52.9      \\
    CBN~\cite{CBN} &ECCV’20    &72.8   & 42.9      \\
    Circle~\cite{Circle_loss} &CVPR’20  &76.3   & -         \\
    SAN~\cite{SAN} &AAAI’20     &79.2   & 55.7      \\
    RGA-SC~\cite{RGA} &CVPR’20  &80.3   & 57.5      \\
    \hline
    TransReID~\cite{transreid} &ICCV’21 &86.2 &69.4 \\    
    DRL-Net~\cite{DRL} &TMM’22  &78.4 & 55.3   \\
    PFD~\cite{PFD} &AAAI’22  &83.8   &64.4   \\
    \hline\hline
    \textbf{\ours} (Ours) & IEEE TIFS & \textbf{88.1}  & \textbf{72.3}  \\
    \hline
    \end{tabular}}
    \label{tab:msmt17}
\end{table}

\subsection{Ablation Study}

We conduct comprehensive ablation studies on Occluded-DukeMTMC to explore the effectiveness of the proposed modules of our \ours.

\textbf{Analysis of Proposed Modules.} In \tablename~\ref{tab:modules}, we evaluate the effectiveness of the Region Generation Module (RGM) and Region Assessment Module (RAM), where
DAI, and IAI are the abbreviation for discrimination-aware indicator and invariance-aware indicator, respectively. 
The baseline utilizes pure ViT pre-trained with CLIP without Overlapping Patches~\cite{transreid} as a feature extractor and is trained under the conventional ID loss and triplet loss. It can achieve 66.6\% Rank-1 and 57.6\% mAP owing to the strong generalization ability of CLIP which already beats most of the SOTA methods.
Next, several observations can be drawn as follows. 
Firstly, When a naive fixed-stripe (Index 1) which shares the same architecture with PCB\cite{PCB} is appended into the baseline (Index 0), the performance is promoted slightly. This is because the fixed stripes are susceptible to misalignment and noise brought from the background or occlusion. By contrast, when adding a single RGM (Index 5) based on the baseline, the performance is improved by +4.0\% and +3.2\% in terms of Rank-1 and mAP, which demonstrates that our RGM is capable of focusing more on extracting discriminative regions while avoiding being affected by noisy information.
Secondly, no matter whether based on the fixed-stripe or the RGM models, further improvement can be achieved by adding DAI or IAI. This implies the strong generalization ability of the DAI and IAI. 
Thirdly, DAI and IAI are complementary to each other, and the combination of them can further improve the RGM results by +1.0\% and +1.6\% in terms of Rank-1 and mAP based on RGM.
Finally, when combining the RGM and RAM (both DAI and IAI) together, \ours achieves the best performance with \textbf{71.6\%} Rank-1 and \textbf{62.4\%} mAP.

\begin{table}[t]
\center
\caption{Components analysis of the proposed \ours on Occluded-DukeMTMC. (DAI, and IAI are the abbreviation for discrimination-aware indicator and invariance-aware indicator, respectively.)}
\setlength{\tabcolsep}{3.mm}
\renewcommand{\arraystretch}{1.1}{
\begin{tabular}{l|ccccc}
\Xhline{1.0pt}
\rowcolor{mygray}&   &\multicolumn{2}{c}{RAM}  &  &  \\ 
\cline{3-4}
\rowcolor{mygray}\multirow{-2}{*}{Type} & \multirow{-2}{*}{Index}  & DAI & IAI & \multirow{-2}{*}{Rank-1} & \multirow{-2}{*}{mAP}  \\ 
\hline
\hline
Baseline &0 & \xmarkg& \xmarkg  &66.6  &  57.6  \\
\hline 
\multirow{4}{*}{Fixed-stripe}& 1 & \xmarkg& \xmarkg &  68.9  &  58.8 \\
&2 & \cmark& \xmarkg  &  69.2  &  59.6  \\
&3 & \xmarkg& \cmark  &  70.0 &  60.2 \\
&4 & \cmark& \cmark   &  70.2  &  61.0 \\
\hline
\multirow{4}{*}{RGM}& 5 & \xmarkg& \xmarkg &70.6 &  60.8 \\
&6 &  \cmark& \xmarkg & 71.0  &  61.4 \\
&7 &  \xmarkg& \cmark & 71.1  &  61.6 \\
&8 &  \cmark& \cmark & 71.6 & 62.4 \\

\hline
\end{tabular}}
\label{tab:modules}
\end{table}

\textbf{Prompt design}.
We compare five prompt designs methods for RGM only to avoid the impact of RAM. The comparison results are
shown in \tablename~\ref{tab:prompt_design}. 
The naive “a photo of a [CLASS].” obtain 57.1\% mAP and simply inserting the word “person” before  “photo” would improve the result. While adding “part of a person” into the naive prompt would
hurt the performance. Therefore, we remove “photo of” and directly utilize “a [CLASS] part of a person.”, which improves the result by +2.4\% mAP. 
Finally, the learnable prompt outperforms all the manually searched
prompts by +1.3\% mAP, clearly showing the generalization power of the learnable prompt.
It is worth noting that hand-crafted templates do not need extra training process or external pseudo parsing segmentation and the result surpasses fixed-stripe based methods by +0.7\% mAP, which verifies the superiority of our design.
\begin{table}[t]
\center
\caption{Performance of \ours with different prompt designs
on Occluded-Duke. [CLASS] represents the class token, and [Learnable
Tokens] represent learnable prompts.}
\setlength{\tabcolsep}{3.mm}
\renewcommand{\arraystretch}{1.1}
{
\begin{tabular}{l|cc}
\Xhline{1.0pt}
\rowcolor{mygray}Prompts   & Rank-1 & mAP  \\ 
\hline
\hline
“a photo of a [CLASS].”  & 67.6 & 57.1  \\
\rowcolor{mygray2}“a person photo of a [CLASS].”  & 68.7 & 58.6  \\
“a photo of a [CLASS] part of a person.”  & 68.1 & 58.2  \\
\rowcolor{mygray2}“a [CLASS] part of a person.”  & 69.2 &59.5\\
“[Learnable Tokens] + [CLASS]”  &70.6  & 60.8  \\
\hline
\end{tabular}}
\label{tab:prompt_design}
\end{table}

\begin{figure*}[t]
	\centering
	\includegraphics[scale=0.56]{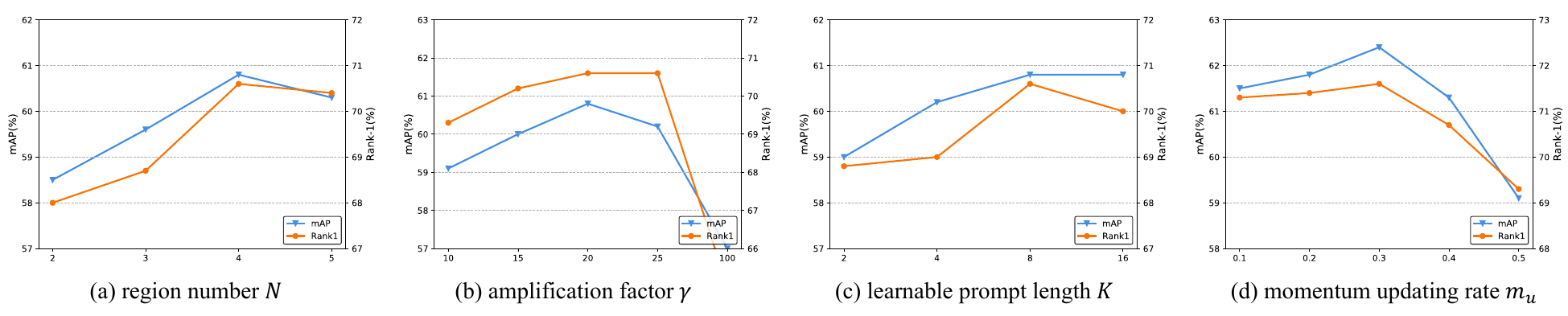}
	\caption{\small{Effects of three hyper-parameters, the region number $n$, amplification factor $\gamma$, learnable prompt length $K$, and momentum updating rate $m_u$ on Occluded-DukeMTMC.}}
	\label{fig:region_number}
\end{figure*}

\textbf{Evaluation of the Region Number $N$.} To explore the influence of the different number of generated regions, we conduct several experiments based on RGM only without RAM and show the results (Rank-1 and mAP) in~\figurename~\ref{fig:region_number} (a). As can be seen, with $N$ increase, the result keeps improving at first and reaches the peak: 70.6\% Rank-1 accuracy and 60.8\% mAP when $N$ arrives 4. The conclusion is consistent with past methods like PCB and VPM. but overly large region numbers consume more computing resources and slow down the inference speed. Thus, we choose region number 4 to achieve a good balance of accuracy and efficiency. The pre-defined text names for different regions are set to 
( ``\texttt{upper body}'', ``\texttt{lower body}'', 2 regions in total),
( ``\texttt{head}'', ``\texttt{upper body}'', and ``\texttt{lower body}'', 3 regions in total),
( ``\texttt{head}'', ``\texttt{upper body}'', ``\texttt{lower body}'', and ``\texttt{foot}'', 4 regions in total),
( ``\texttt{head}'', ``\texttt{upper body}'', ``\texttt{lower body}'', ``\texttt{foot}'', and ``\texttt{bags}'', 5 regions in total).

\textbf{Analysis of $\gamma$ of RGM.} We conduct experiments on Occluded-DukeMTMC to analyze the impact of $\gamma$ based on RGM only, which denotes the amplification factor for similarity measurement for visual and textual features. As shown in the~\figurename~\ref{fig:region_number} (b), too small $\gamma$ leads to poor performance, because the similarity is too small to separate these regions apart. Too large $\gamma$ like 100 may result in unstable training and loss explosion, which degrades the performance significantly. We set $\gamma$ to 20 in our experiments, which achieves the best result with robustness. 

\textbf{Impact of $K$ in the learnable prompts.} To investigate the impact of $k$, we perform experiments based on RGM only and show the results in the~\figurename~\ref{fig:region_number} (c), our model is not sensitive to the parameter $K$, and the curve is relatively flat. Too small $K$ results in poor performance, because there are not enough context parameters to mine ReID discriminative information. Too large $K$ may result in the redundancy of sample varieties and raise the risk of overfitting, which degrades the performance marginally. Therefore, we set $K$ to 8 in our experiments, which achieves robust performance and keeps efficient.

\textbf{Influence of $m_u$.} $m_u$ represents the momentum updating rate used for the invariance-aware indicator in the RAM. As shown in~\figurename~\ref{fig:region_number} (d), $m_u=0.3$ achieves the best performance and is chosen as our default setting. We analyze the performance and conclude that too-small $m_u$ leads to the slow updating of the memory bank feature and too-large $m_u$ results in the drastic updating of the memory bank feature. $m_u=0.3$ achieves a good balance between the robustness and up-to-dateness of the memory bank.

\begin{figure*}[t]
	\includegraphics[width=1.0\textwidth]{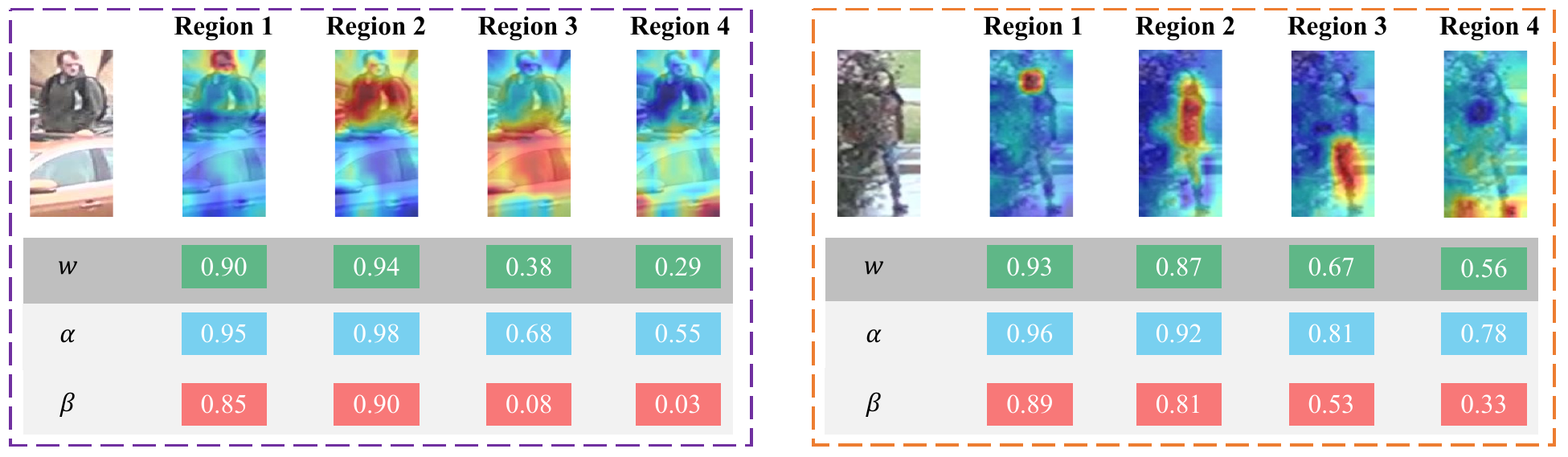}
	\caption{Visualization of our \ours under heavy occlusions. $\alpha$, $\beta$ and $w$ are discrimination score, invariance score, and final confidence score, respectively. } 
	\label{fig:visualization}
\end{figure*}

\begin{figure*}[t]
	\includegraphics[width=1.0\textwidth]{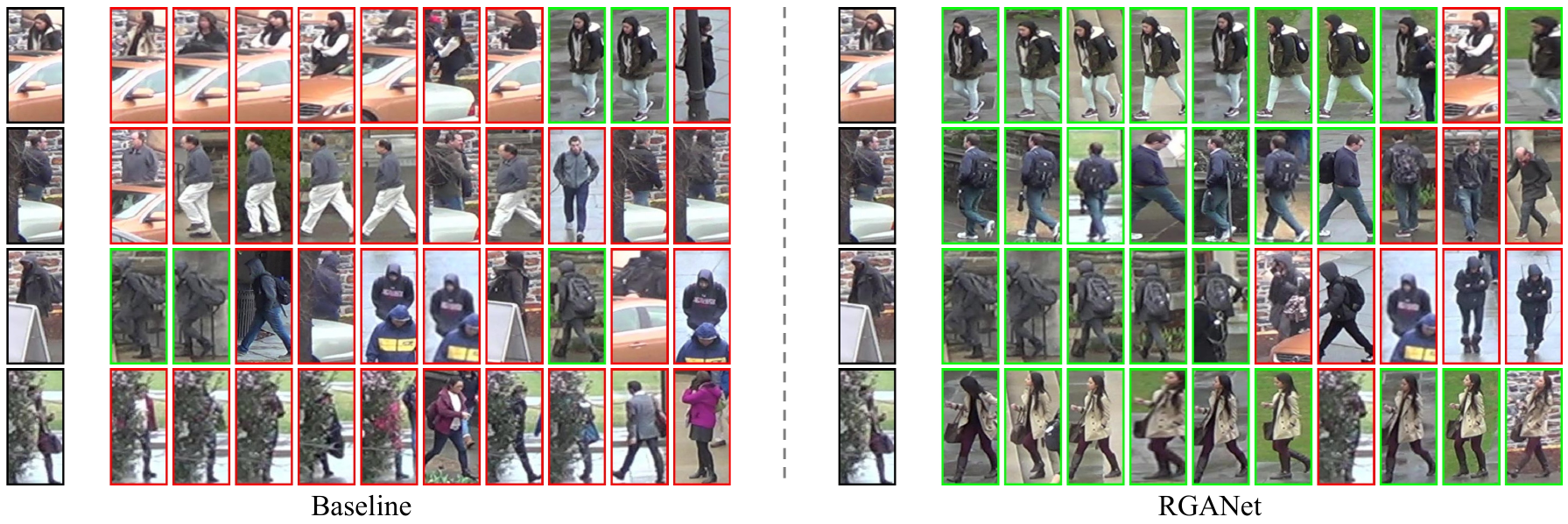}
	\caption{Person image ranking of our baseline and the proposed \ours are presented on Occluded-DukeMTMC dataset. 
 Green and red rectangles represent right and wrong retrieval results for a given query image, respectively.} 
	\label{fig:visualization_rank}
\end{figure*}

\subsection{Visualization of \ours under Heavy Occlusions}
We visualize the generated regions and their assigned confidence values in \figurename~\ref{fig:visualization}. For clear explanation, DAI, and IAI are the abbreviations for discrimination-aware indicator, invariance-aware indicator. $\alpha$, $\beta$ and $w$ are discrimination score, invariance score, and final confidence score, respectively. 
Take \figurename~\ref{fig:visualization} (a) for example,
the person's lower body and feet are heavily occluded by the occlusions (car), and the RGM has the capability of detecting the approximate positions of occluded regions, \ie, Region 3 and 4, which are useless for discriminative representation learning.
The DAI wrongly assigns Regions 3 and 4 with higher scores, because it can't tell the difference between foreground and background in Regions 3 and 4. The inherent reason is that this man always comes with the car. While the IAI utilizes the auxiliary information from the memory bank to resort to the invariance-level information (complete head, upper body, lower body, and foot). Specifically, Regions 3 and 4 are not invariant features in the memory bank, therefore the IAI values of these regions are lower. $w$ is the mean of the $\alpha$ and the $\beta$, which combines their respective advantages. The experiment results further show that jointly considering the discrimination and the invariance is more effective for person ReID.

\subsection{Qualitative Retrieval Results}
To further show the effectiveness of our proposed \ours, we provide qualitative retrieval results in \figurename~\ref{fig:visualization_rank}.
The provided image samples are full of different kinds of occlusions (cars, boards, trees).
For each occluded person image from query set, there are the top 10 gallery images that are retrieved by the baseline and our \ours, respectively. 
Green and red rectangles represent right and wrong retrieval results for a given query image, respectively.
It can be seen that our \ours is capable of alleviating the occlusions and retrieving the corresponding the same identity images correctly. By contrast, the baseline is susceptible to occlusions and outputs a lot of wrong person images. Therefore, our \ours can suppress background and
occluded parts that are detrimental to the ReID performance.

\subsection{Computational Complexity}
\begin{table}[t]
\center
\caption{Computational complexity Comparison with state-of-the-art models
on Occluded-Duke dataset.}
\setlength{\tabcolsep}{3.mm}
\renewcommand{\arraystretch}{1.1}
{
\begin{tabular}{lcc|cc}
\Xhline{1.0pt} 
\rowcolor{mygray}Methods   & FPS   &\#Params.   & Rank-1 & mAP  \\ 
\hline
\hline
PGFA~\cite{PGFA}   & 156.4 &  115.4M & 51.4 & 37.3  \\
HOReID~\cite{HOReID} & 62.9 & 117.6M & 55.1 & 43.8\\
PAT~\cite{PAT} &284.6  & 67.5M &64.5  &53.6 \\
DPM~\cite{DPM} &98.3  & 146.2M & 71.4 &61.8 \\

\hline
\textbf{\ours} (Ours) &  213.9 & 93.5M & 71.6 &62.4 \\
\hline
\end{tabular}}
\label{tab:complexity}
\end{table}

We conduct further experiments in \tablename~\ref{tab:complexity} to show that our \ours not only achieves superior results but also has advantages on inference speed and model complexity. We compare recent popular occluded ReID methods: PGFA~\cite{PGFA}, HOReID~\cite{HOReID}, and DPM~\cite{DPM}.
To ensure a fair comparison, 
inference batch size is all set to 128. All the experiments are conducted with one Nvidia RTX TITAN GPU using the PyTorch toolbox. The inference time contains both the feature extraction of images and the distance calculation between query and galley images.
From the \tablename~\ref{tab:complexity}, we can see that our \ours has relatively small model parameters because we just add several convolution or linear layers apart from our ViT baseline. Besides, the inference speed of \ours is significantly faster than external tool based methods (PGFA~\cite{PGFA} and HOReID~\cite{HOReID}) because we get rid of external tools in the inference stage. Although DPM~\cite{DPM} is also a transformer based method, it uses the operation of Overlapping Patches and multiple $3\times 3$ convolutional layers to extract mask, which will greatly increase inference time. While PAT has a slight advantage in terms of smaller parameters and faster inference speed, our method significantly outperforms it in terms of performance.
The comparisons demonstrate that our proposed \ours is both compact and efficient.

%% file: 5_conclusion.tex
\section{Conclusion}
In this paper, to address the challenges of misalignment, background variations, and occlusions in person ReID task, we propose an effective and efficient \textbf{R}egion \textbf{G}eneration and \textbf{A}ssessment \textbf{Net}work (\textbf{RGANet}), which takes advantage of CLIP to capture the discriminative and invariant region features. Firstly, Region Generation Module (RGM) is designed to automatically search and locate the more discriminative regions. Meanwhile, to obtain reasonable contributions of generated regions, Region Assessment Module (RAM) is proposed to assess each region by a discrimination-aware indicator and an invariance-aware indicator, which are proved to be complementary with each other from both qualitative and quantitative perspectives. Extensive experiments on occluded, partial, and holistic datasets consistently demonstrate the superiority of the proposed approach.